\documentclass{article}
\usepackage{stywhispers,amsmath,epsfig}

\title{NATIONAL-SCALE CROP TYPE MAPPING IN MOROCCO USING GOOGLE ALPHAEARTH FOUNDATIONS EMBEDDINGS AND THE FIELDS OF THE WORLD FIELD BOUNDARY MAP}
\name{Walid Elbarz$^{1}$\thanks{Corresponding author: walid.ELBARZ@um6p.ma}\thanks{Presented at GeoAI4SD 2026 (online, July 9, 2026).}, Mohamed Bourriz$^{1,2,3}$, Vincent Simonneaux$^4$, Hamd Ait Abdelali$^1$, François Bourzeix$^1$}
\address{
$^1$Analytics Lab (A-Lab), UM6P, Rabat 11103, Morocco\\
$^2$Center for Remote Sensing Applications (CRSA), UM6P, Ben Guerir 43150, Morocco\\
$^3$Friedrich Schiller University Jena, Department of Geography, Jena 07743, Germany\\
$^{4}$Centre d'Études Spatiales de la Biosphère (CESBIO), Université de Toulouse, Toulouse, France
}
\begin{document}
%
\maketitle
\begin{abstract}
Recent geospatial foundation models have introduced a new paradigm for large-scale Earth observation by providing pretrained embedding representations that can be directly exploited for downstream tasks. In this work, we investigate the use of Google AlphaEarth Foundations embeddings for national-scale crop type mapping in Morocco. The proposed workflow operates directly on publicly available 64-dimensional pretrained embeddings using lightweight downstream classifiers, eliminating the need to process raw satellite imagery during model training and inference. The approach targets three crop categories, namely winter cereals, tree crops, and other crops. For nationwide inference, the Fields of the World global field boundary map is used to constrain predictions to agricultural parcels, enabling efficient large-scale mapping. Experimental evaluation on an independent test set over ten repeated runs achieved an average classification accuracy of $(87.17 \pm 0.42)\%$. Nationwide inference over Morocco was completed in approximately five hours, with the dominant computational cost arising from cloud-based retrieval of the embeddings rather than model execution. These results demonstrate the potential of pretrained embedding representations for computationally efficient national-scale crop type mapping.
\end{abstract}
\section{Introduction}
\label{sec:intro}

Remote sensing has become an indispensable technology for large-scale Earth observation, enabling continuous monitoring of environmental, agricultural, and ecological systems. Among its numerous applications, crop type mapping plays a central role in precision agriculture, agricultural statistics, food security assessment, and land-use management. Compared with traditional field surveys, satellite remote sensing provides a cost-effective and scalable alternative for producing crop maps over large geographic regions. Recent advances in artificial intelligence, machine learning, and big-data analytics have further accelerated this transition, enabling increasingly accurate crop classification from multi-temporal satellite imagery \cite{Bourriz2025}.

Despite these advances, developing reliable crop mapping models remains challenging in regions where annotated agricultural data are scarce. Many countries, particularly in the Global South, lack the extensive ground-truth datasets required to train modern deep learning models, resulting in limited generalization and reduced mapping accuracy. Collecting field observations at national scale is both time-consuming and financially expensive, creating a major bottleneck for operational crop monitoring systems.

The emergence of geospatial foundation models offers a promising paradigm for addressing this limitation. Large-scale models such as TerraMind \cite{terramind}, SpectralEarth-FM \cite{spectral_earth}, and related foundation models leverage self-supervised pretraining on massive volumes of Earth observation data to learn generic representations that can be transferred to downstream remote sensing tasks with minimal labeled data. These models have demonstrated strong performance across a wide range of benchmarks, suggesting that pretrained representations can substantially reduce the dependence on extensive local annotations.

More recently, a complementary paradigm has emerged through the use of precomputed embedding representations \cite{pool}. Instead of processing raw satellite imagery during downstream inference, foundation models encode multi-temporal and multimodal observations into compact, high-dimensional feature vectors that capture both spectral and temporal information. This approach significantly reduces computational requirements while preserving the information necessary for classification. The concept was first demonstrated on a global scale by the Presto foundation model \cite{presto} in the WorldCereal project and has since been further advanced by Google's AlphaEarth Foundations (AEF) embeddings \cite{AEF} and the Tessera foundation modelc\cite{Tessera}. These embedding-based representations enable efficient large-scale analysis by replacing computationally intensive feature extraction with reusable latent representations. Recent studies have demonstrated that embedding-based workflows can not only improve classification performance but also substantially reduce storage, preprocessing, and inference costs, making national-scale mapping considerably more practical.

The availability of global foundation models, together with increasingly accessible planetary-scale Earth observation datasets, opens new opportunities for operational crop mapping in data-scarce regions. Rather than training complex deep learning architectures from scratch, downstream classifiers can be trained directly on foundation model embeddings, enabling efficient deployment while requiring only a limited amount of local training data.

In this context, this paper investigates the capability of AEF embeddings for large-scale crop type mapping across Morocco. Specifically, we evaluate the effectiveness of pretrained embedding representations for national-scale crop classification and demonstrate a computationally efficient workflow that substantially reduces the cost of feature extraction and model deployment while maintaining competitive mapping accuracy.

\section{Study Area}
\begin{figure}[!t]
    \centering
    \includegraphics[width=\columnwidth]{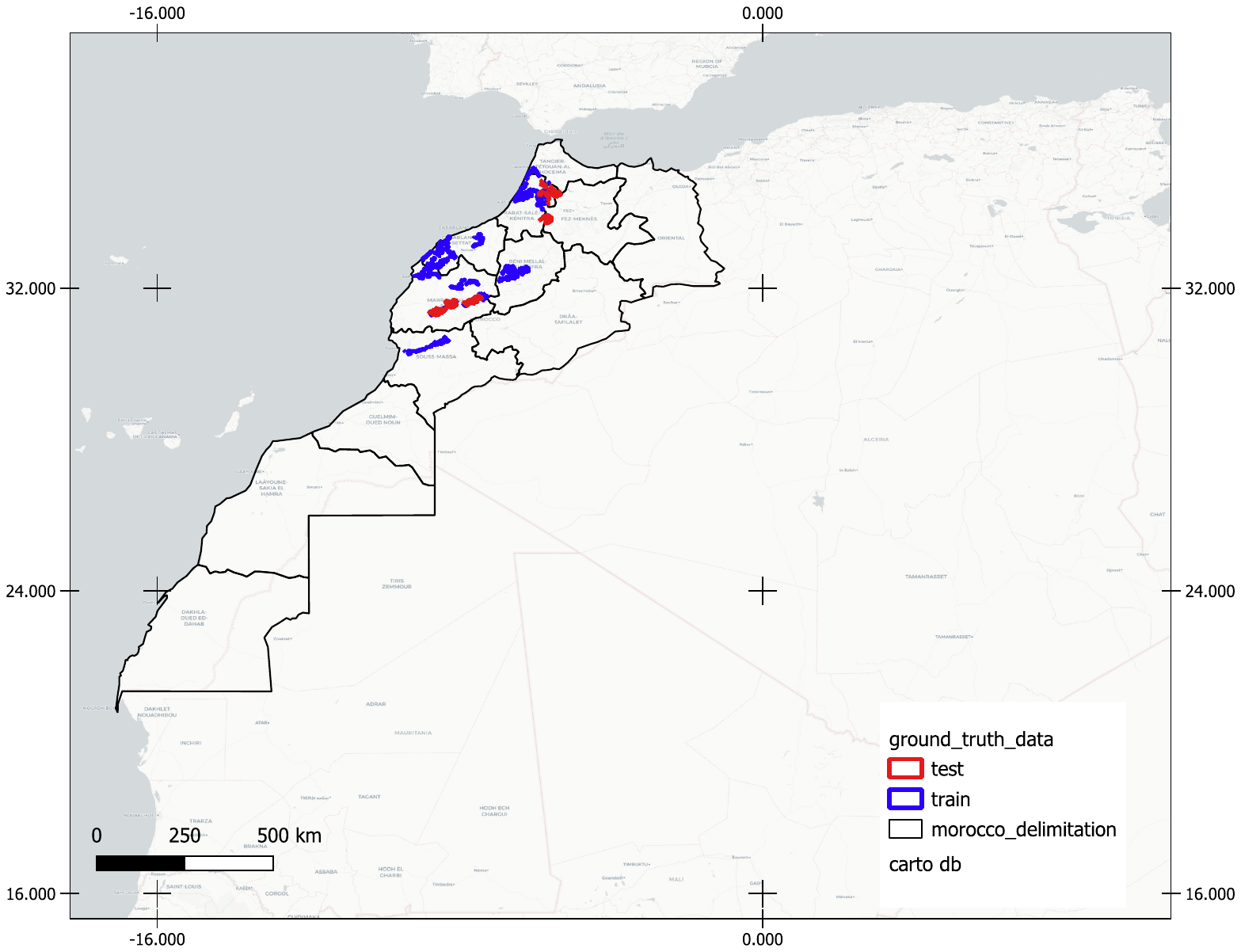}
    \caption{Study area showing the agricultural regions considered in this work and the distribution of the reference data used for model development and evaluation.}
    \label{fig:study_area}
\end{figure}

\label{sec:format}

The study covers the major agricultural regions of Morocco, including Souss-Massa, Marrakech-Safi, Béni Mellal-Khénifra, Casablanca-Settat, Rabat-Salé-Kénitra, and Tanger-Tétouan-Al Hoceïma (Figure~\ref{fig:study_area}). These regions encompass a broad range of agro-climatic conditions, agricultural systems, and cropping practices, making them representative of Morocco's agricultural landscape. The geographical diversity of the study area provides a suitable benchmark for evaluating the robustness and generalization capability of foundation model embeddings for national-scale crop type mapping.

Ground-truth data were compiled from the publicly available Moroccan agricultural database, which provides field-level crop annotations across the selected regions \cite{grich}. To further evaluate the generalization capability of the proposed workflow, an independent validation dataset was assembled using reference data collected during the 2025 agricultural season together with field observations acquired in the Tensift region between 2018 and 2024.
The complete reference dataset contains \textbf{13,844} agricultural polygons distributed among three crop classes: winter cereals, tree crops, and other crops. Figure~\ref{fig:study_area} illustrates the geographical distribution of the study regions and the corresponding ground-truth data.

\section{Remote Sensing Data}
\label{sec:data}

This study relies exclusively on the Google AlphaEarth Foundations (AEF) embedding dataset \cite{AEF}. The AEF is a geospatial foundation model that learns compact latent representations from large volumes of multi-temporal and multimodal satellite observations through self-supervised learning. Instead of operating directly on raw satellite imagery, the model encodes each observation into a fixed-length embedding vector that captures its spectral, temporal, and contextual characteristics, allowing the learned representations to be transferred efficiently to downstream Earth observation tasks.

The publicly available AEF embeddings, distributed through an open AWS S3 bucket, are provided at a spatial resolution of 10~m. Each observation is represented by a 64-dimensional quantized embedding vector, which is dequantized prior to model training. Throughout this work, these embeddings constitute the only input features used for crop classification.

The proposed workflow aims to classify agricultural observations into three general crop categories: winter cereals, tree crops (arboriculture), and other crops.
To facilitate nationwide crop mapping, the Fields of the World (FTW) global field boundary map \cite{ftw} is employed during inference. The FTW parcel layer is used as a spatial mask restricting inference to agricultural areas, thereby avoiding predictions over non-agricultural land cover. For every agricultural parcel, the corresponding AEF embedding is retrieved and classified into one of the three target crop categories. This strategy substantially reduces the computational cost of nationwide inference while preserving complete coverage of cultivated land. Although the FTW map may require local refinement, it provides an excellent starting point for large-scale operational crop mapping at a fraction of the computational cost.

\begin{figure*}[!t]
    \centering
    \includegraphics[width=\textwidth]{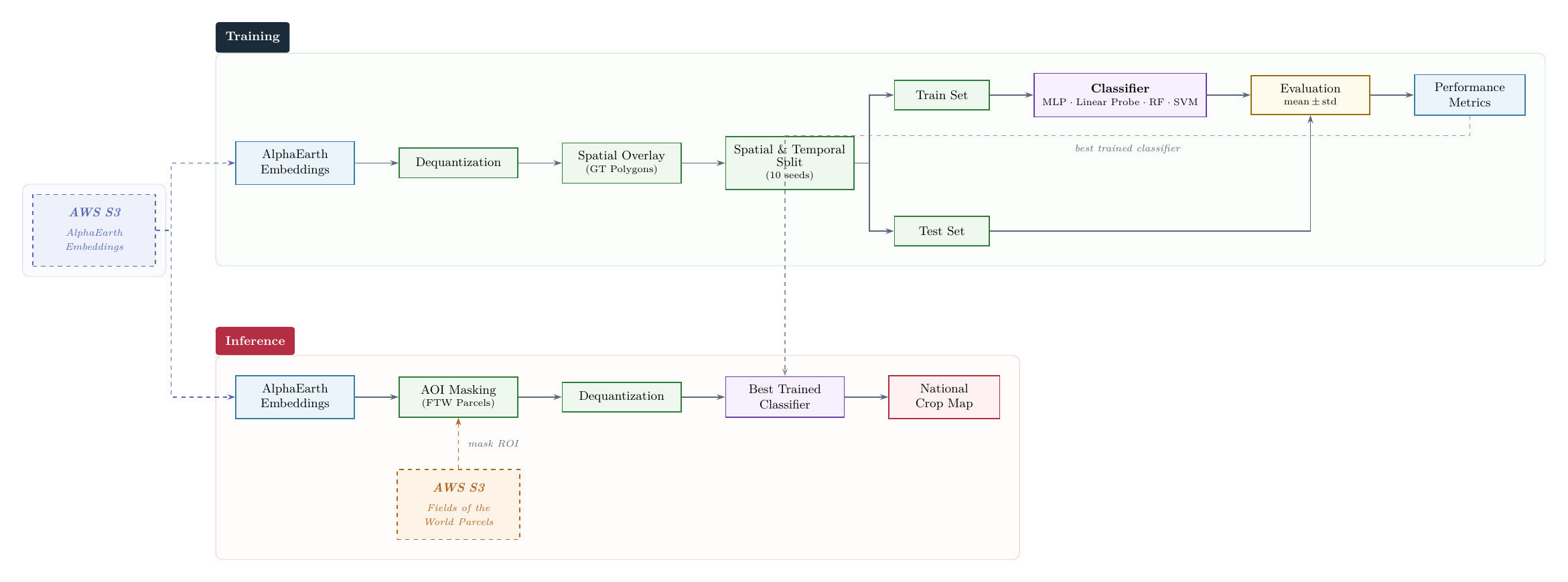}
    \caption{Overview of the proposed workflow. Publicly available AEF embeddings are first dequantized and combined with ground-truth reference polygons to construct the training dataset. Lightweight downstream classifiers are then trained and evaluated on an independent test set. During nationwide inference, the selected classifier is applied to the AEF embeddings within the Fields of the World field boundary map to produce the final nationwide crop type map.}
    \label{fig:workflow}
\end{figure*}

\section{Training and Inference Methodology}
\label{sec:methodology}

The complete workflow adopted in this study is illustrated in Figure~\ref{fig:workflow}. The proposed methodology consists of four main stages: embedding preprocessing, training dataset generation, downstream model training and evaluation, and national-scale crop mapping.

\subsection{Embedding Preprocessing}

The AEF embeddings are distributed as quantized integer values. Prior to model training, each embedding vector is converted to floating-point values using the dequantization procedure proposed by the AEF authors:

\begin{equation}
v_d=
\left(
\frac{v_q}{127.5}
\right)^2
\times
\mathrm{sign}(v_q),
\label{eq:dequantization}
\end{equation}

where $v_q$ denotes the quantized embedding value and $v_d$ its corresponding dequantized representation.

Following dequantization, every sample is represented by a 64-dimensional embedding vector with values approximately ranging from $-0.4764$ to $0.4764$.

\subsection{Training Dataset Generation}
 Reference crop polygons were spatially overlaid with the AEF embedding dataset in order to associate each labeled observation with its corresponding embedding vector. For every reference sample, the extracted information includes the embedding vector, crop label, acquisition year, and additional metadata used during model development.

After preprocessing, the final training pool contains 71,062 labeled samples distributed among the three crop classes. This increase relative to the number of reference polygons results from sampling multiple embedding pixels within individual agricultural polygons. An additional independent test dataset containing 34,565 samples was isolated prior to model development and remained completely untouched during training, model selection, and hyperparameter optimization.

\subsection{Model Training and Spatial Validation}

 The proposed workflow adopts an embedding-based downstream classification paradigm.

Four downstream classifiers are evaluated throughout this study:

\begin{itemize}
\item Linear Probe;
\item Multi-Layer Perceptron (MLP);
\item Random Forest;
\item Support Vector Machine (SVM).
\end{itemize}

To obtain a robust estimate of model performance, each classifier is trained and evaluated over ten independent random seeds. The best-performing classifier is subsequently evaluated on the frozen independent test dataset, which was isolated prior to model development and remained completely untouched throughout training and hyperparameter optimization, in order to assess transferability to previously unseen geographical regions.

\begin{figure*}[!t]
    \centering
    \includegraphics[width=\textwidth]{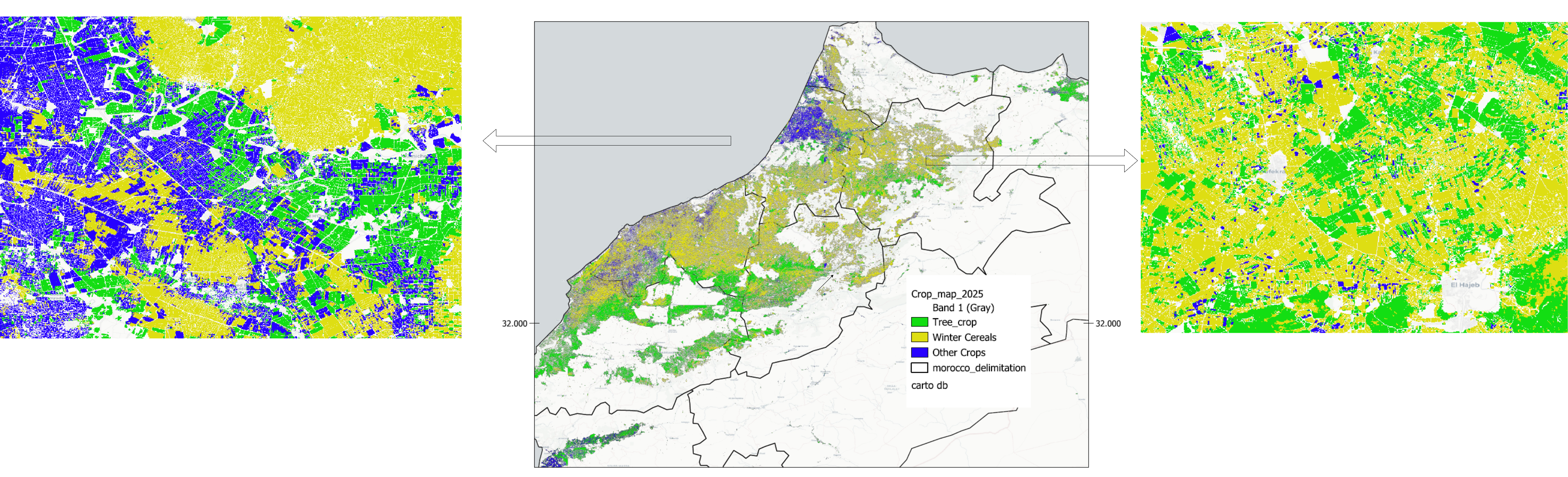}
    \caption{Examples of the national crop type map produced using the proposed workflow. (On the left) Agricultural landscape in the Gharb region. (On the right) Agricultural landscape in the El Hajeb region.}
    \label{fig:inference_examples}
\end{figure*}

\subsection{National-Scale Inference}

Following model selection, the best-performing classifier is deployed to generate a national crop type map over Morocco.

Instead of performing exhaustive pixel-wise prediction across the entire country, inference is restricted to agricultural parcels extracted from the 2024 Fields of the World (FTW) dataset. For each agricultural parcel, the corresponding AEF embedding is retrieved and processed by the trained classifier to predict one of the three target crop classes.

This parcel-guided inference strategy significantly reduces computational requirements while enabling efficient nationwide crop mapping using only pretrained embedding representations. Figure~\ref{fig:workflow} summarizes the complete processing pipeline from AEF embeddings to the final crop type map.

\section{Results}
\label{sec:results}
\subsection{Classification Performance}

The classification performance of the four evaluated downstream models on the independent test set is summarized in Table~\ref{tab:classification_results}. Each model was trained and evaluated over ten independent random seeds; the reported accuracy corresponds to the mean and standard deviation computed across runs.

\begin{table}[!t]
\renewcommand{\arraystretch}{1.3}
\caption{Test accuracy (\%) on the independent test set for each downstream classifier (mean $\pm$ standard deviation over 10 runs).}
\label{tab:classification_results}
\centering
\begin{tabular}{lc}
\hline
\textbf{Model} & \textbf{Test Accuracy (\%)} \\
\hline
Linear Probe  & $86.66 \pm 0.04$ \\
MLP           & $\mathbf{87.17 \pm 0.42}$ \\
Random Forest & $86.88 \pm 0.08$ \\
SVM           & $86.08 \pm 0.24$ \\
\hline
\end{tabular}
\end{table}

The results demonstrate that AEF contain highly discriminative information for agricultural crop classification. Despite relying exclusively on lightweight downstream classifiers, the learned representations achieve competitive and consistent performance across the three target crop categories. The MLP achieves the highest test accuracy at $(87.17 \pm 0.42)\%$, closely followed by Random Forest at $(86.88 \pm 0.08)\%$ and Linear Probe at $(86.66 \pm 0.04)\%$. All models yield test accuracies in the range of $86$--$87\%$ on the independent test set, which consists exclusively of samples from regions excluded from model development, confirming the transferability of the AEF embeddings to previously unseen agricultural areas. Furthermore, the relatively small performance differences between the evaluated classifiers indicate that the pretrained embeddings already provide highly discriminative representations.
\subsection{National-Scale Crop Mapping}

The selected MLP classifier is subsequently applied to the complete 2024 Fields of the World agricultural parcel layer to generate a national crop type map of Morocco. Nationwide inference was completed in approximately five hours, with the dominant computational cost arising from cloud-based retrieval of the AEF embeddings rather than model execution itself. This result highlights the practical scalability of the proposed embedding-based approach for operational crop monitoring at national scale.

Figure~\ref{fig:inference_examples} illustrates representative examples of the produced national crop type map at regional scale.

\section{Conclusion}
\label{sec:conclusion}

This paper presented a computationally efficient workflow for national-scale crop type mapping in Morocco using Google AlphaEarth Foundations (AEF) pretrained embeddings. By operating exclusively on publicly available 64-dimensional embedding representations and lightweight downstream classifiers, the proposed approach eliminates the need for raw satellite imagery processing during both training and inference, substantially reducing the computational burden associated with large-scale crop mapping.

Experimental results on an independent test set demonstrated that the MLP classifier achieves the best performance, with a test accuracy of $(87.17 \pm 0.42)\%$ over ten random seeds, confirming the spatial transferability of the AEF embeddings to previously unseen agricultural regions. Nationwide inference over Morocco, guided by the Fields of the World field boundary map, was completed in approximately five hours, with cloud-based embedding retrieval identified as the dominant computational bottleneck.

These findings demonstrate that foundation model embeddings offer a viable and scalable solution for crop type mapping in data-scarce regions, where collecting extensive ground-truth annotations is impractical. The combination of globally pretrained representations with the FTW agricultural parcel layer provides a powerful and cost-effective framework for operational crop monitoring at a national scale.

Future work will focus on incorporating additional crop categories, evaluating the generalization of the workflow across different agro-climatic contexts, and exploring multi-temporal embedding aggregation strategies to further improve classification accuracy.

\section*{Acknowledgements}

The authors would like to thank Taylor Geospatial for publishing the Google AlphaEarth Foundations embeddings on Source Cooperative, and Source Cooperative, a non-profit open data publishing utility created by Radiant Earth, for providing the infrastructure that made these embeddings openly accessible to the research community. This work was also conducted within the Yield Gap project, supported by the OCP Foundation–UM6P agreement.

\bibliographystyle{IEEEbib}
\bibliography{refs}

@article{Bourriz2025,
  title = {Integration of Hyperspectral Imaging and AI Techniques for Crop Type Mapping: Present Status,  Trends,  and Challenges},
  volume = {17},
  ISSN = {2072-4292},
  url = {http://dx.doi.org/10.3390/rs17091574},
  DOI = {10.3390/rs17091574},
  number = {9},
  journal = {Remote Sensing},
  publisher = {MDPI AG},
  author = {Bourriz,  Mohamed and Hajji,  Hicham and Laamrani,  Ahmed and Elbouanani,  Nadir and Abdelali,  Hamd Ait and Bourzeix,  Fran\c{c}ois and El-Battay,  Ali and Amazirh,  Abdelhakim and Chehbouni,  Abdelghani},
  year = {2025},
  month = Apr,
  pages = {1574}
}

@misc{terramind,
  doi = {10.48550/ARXIV.2504.11171},
  url = {https://arxiv.org/abs/2504.11171},
  author = {Jakubik,  Johannes and Yang,  Felix and Blumenstiel,  Benedikt and Scheurer,  Erik and Sedona,  Rocco and Maurogiovanni,  Stefano and Bosmans,  Jente and Dionelis,  Nikolaos and Marsocci,  Valerio and Kopp,  Niklas and Ramachandran,  Rahul and Fraccaro,  Paolo and Brunschwiler,  Thomas and Cavallaro,  Gabriele and Bernabe-Moreno,  Juan and Longépé,  Nicolas},
  title = {TerraMind: Large-Scale Generative Multimodality for Earth Observation},
  publisher = {arXiv},
  year = {2025},
  copyright = {arXiv.org perpetual,  non-exclusive license}
}

@misc{spectral_earth,
  doi = {10.48550/ARXIV.2605.21075},
  url = {https://arxiv.org/abs/2605.21075},
  author = {Braham,  Nassim Ait Ali and Banze,  Aaron and Albrecht,  Conrad M. and Mairal,  Julien and Chanussot,  Jocelyn and Zhu,  Xiao Xiang},
  title = {SpectralEarth-FM: Bringing Hyperspectral Imagery into Multimodal Earth Observation Pretraining},
  publisher = {arXiv},
  year = {2026},
  copyright = {Creative Commons Attribution 4.0 International}
}

@misc{pool,
  doi = {10.48550/ARXIV.2603.02080},
  url = {https://arxiv.org/abs/2603.02080},
  author = {Corley,  Isaac and Robinson,  Caleb and Becker-Reshef,  Inbal and Ferres,  Juan M. Lavista},
  title = {From Pixels to Patches: Pooling Strategies for Earth Embeddings},
  publisher = {arXiv},
  year = {2026},
  copyright = {Creative Commons Attribution 4.0 International}
}

@misc{presto,
  doi = {10.48550/ARXIV.2304.14065},
  url = {https://arxiv.org/abs/2304.14065},
  author = {Tseng,  Gabriel and Cartuyvels,  Ruben and Zvonkov,  Ivan and Purohit,  Mirali and Rolnick,  David and Kerner,  Hannah},
  title = {Lightweight,  Pre-trained Transformers for Remote Sensing Timeseries},
  publisher = {arXiv},
  year = {2023},
  copyright = {arXiv.org perpetual,  non-exclusive license}
}

@misc{AEF,
  doi = {10.48550/ARXIV.2507.22291},
  url = {https://arxiv.org/abs/2507.22291},
  author = {Brown,  Christopher F. and Kazmierski,  Michal R. and Pasquarella,  Valerie J. and Rucklidge,  William J. and Samsikova,  Masha and Zhang,  Chenhui and Shelhamer,  Evan and Lahera,  Estefania and Wiles,  Olivia and Ilyushchenko,  Simon and Gorelick,  Noel and Zhang,  Lihui Lydia and Alj,  Sophia and Schechter,  Emily and Askay,  Sean and Guinan,  Oliver and Moore,  Rebecca and Boukouvalas,  Alexis and Kohli,  Pushmeet},
  title = {AlphaEarth Foundations: An embedding field model for accurate and efficient global mapping from sparse label data},
  publisher = {arXiv},
  year = {2025},
  copyright = {Creative Commons Attribution 4.0 International}
}

@misc{ftw,
  doi = {10.48550/ARXIV.2605.11055},
  url = {https://arxiv.org/abs/2605.11055},
  author = {Robinson,  Caleb and Muhawenayo,  Gedeon and Khanal,  Subash and Fang,  Zhanpei and Corley,  Isaac and Tárano,  Ana M. and Estes,  Lyndon and Marcus,  Jennifer and Jacobs,  Nathan and Kerner,  Hannah and Becker-Reshef,  Inbal and Ferres,  Juan M. Lavista},
  title = {The first global agricultural field boundary map at 10m resolution},
  publisher = {arXiv},
  year = {2026},
  copyright = {Creative Commons Attribution 4.0 International}
}

@misc{grich,
  doi = {10.6084/M9.FIGSHARE.28486022.V2},
  url = {https://figshare.com/articles/dataset/_b_Crop_and_irrigation_type_dataset_for_Moroccan_agricultural_regions_b_b_A_ground-truth_resource_for_earth_observation_validation_b_/28486022/2},
  author = {OUASSANOUAN,  Youness and Elfarkh,  Jamal and Grich,  Said and Liblab,  Abderrahman and Chehbouni,  Abdelghani},
  title = {Crop and irrigation type dataset for Moroccan agricultural regions: A ground-truth resource for earth observation validation},
  publisher = {figshare},
  year = {2025},
  copyright = {Creative Commons Attribution 4.0 International}
}

@misc{Tessera,
  doi = {10.48550/ARXIV.2506.20380},
  url = {https://arxiv.org/abs/2506.20380},
  author = {Feng,  Zhengpeng and Atzberger,  Clement and Jaffer,  Sadiq and Knezevic,  Jovana and Sormunen,  Silja and Young,  Robin and Lisaius,  Madeline C. and Immitzer,  Markus and Jackson,  Toby and Ball,  James and Coomes,  David A. and Madhavapeddy,  Anil and Blake,  Andrew and Keshav,  Srinivasan},
  title = {TESSERA: Temporal Embeddings of Surface Spectra for Earth Representation and Analysis},
  publisher = {arXiv},
  year = {2025},
  copyright = {Creative Commons Attribution 4.0 International}
}

\end{document}